%% file: 0-main.tex
\pdfoutput=1

\documentclass[11pt]{article}

\usepackage{ACL2023}

\usepackage{times}
\usepackage{latexsym}

\usepackage[T1]{fontenc}

\usepackage[utf8]{inputenc}

\usepackage{microtype}

\usepackage{inconsolata}

\usepackage{url}
\usepackage{booktabs}
\usepackage{multirow}
\usepackage{amsmath}
\usepackage{dsfont}
\usepackage{algorithm}
\usepackage[noend]{algpseudocode}
\usepackage{pifont}

\usepackage{inconsolata}

\usepackage{blindtext}
\usepackage{graphicx}
\usepackage{capt-of}
\usepackage{booktabs}
\usepackage{varwidth}
\usepackage{multirow}
\usepackage{xspace,mfirstuc,tabulary}

\newcolumntype{L}[1]{>{\raggedright\let\newline\\\arraybackslash\hspace{0pt}}m{#1}}
\newcolumntype{C}[1]{>{\centering\let\newline\\\arraybackslash\hspace{0pt}}m{#1}}

\usepackage{bm}
\usepackage{color}
\usepackage{graphicx}
\usepackage[toc,page]{appendix}
\usepackage{makecell}
\usepackage{boldline}

\usepackage{colortbl}
\usepackage{tabstackengine}
\usepackage{tikz}
\usepackage{relsize}
\usepackage{tablefootnote}
\usepackage{silence}
\usepackage{bbm}

\usepackage{tikz}
\usepackage{soul}

\usepackage{fdsymbol}

\definecolor{lightgray}{gray}{0.9}
\colorlet{soulgreen}{green!30}
\definecolor{red}{HTML}{FF0000}
\definecolor{blue}{HTML}{0000FF}
\definecolor{darkgreen}{HTML}{228B22}
\definecolor{dblue}{HTML}{007FFF}

\usepackage{booktabs,arydshln}

\usepackage{colortbl}
\usepackage{tabstackengine}
\usepackage{tikz}
\usepackage{relsize}
\usepackage{microtype}
\definecolor{magenta}{HTML}{F3DFF1}
\definecolor{hlgreen}{HTML}{ccfcc4}
\definecolor{figblue}{HTML}{e7f2fe}

\DeclareRobustCommand{\greenbox}[1]{\setlength{\fboxsep}{1.0pt}\colorbox{green!25}{#1}}
\DeclareRobustCommand{\redbox}[1]{\setlength{\fboxsep}{1.0pt}\colorbox{red!15}{#1}}

\usepackage{pifont}
\newcommand{\xmark}{\textcolor{red}{\ding{55}}}
\newcommand{\cmark}{\textcolor{darkgreen}{\ding{51}}}

\usepackage{color}
\usepackage{soul}
\makeatletter
\newcommand*\iftodonotes{\if@todonotes@disabled\expandafter\@secondoftwo\else\expandafter\@firstoftwo\fi}  
\makeatother

\usepackage{subfig}
\usepackage{pifont}

\title{Python Code Generation by Asking Clarification Questions}

\author{\textbf{Haau-Sing Li}$^{1}$, \textbf{Mohsen Mesgar}$^{2}$\thanks{\hspace{2mm}Work done while being a postdoc at UKP Lab.}, \textbf{André F. T. Martins}$^{3,4,5}$, \textbf{Iryna Gurevych}$^{1}$
        \\ ${}^{1}$Ubiquitous Knowledge Processing Lab (UKP Lab)\\ Department of Computer Science and Hessian Center for AI (hessian.AI), TU Darmstadt
        \\
        ${}^{2}$Bosch Center for Artificial Intelligence, Renningen, Germany
        \\ ${}^{3}$Instituto Superior Técnico and LUMLIS (Lisbon ELLIS Unit)
        \\ 
        ${}^{4}$Instituto de Telecomunicações, Lisbon, Portugal
        \quad       
        ${}^{5}$Unbabel
        \\ \texttt{hli@ukp.tu-darmstadt.de}
        }

\begin{document}
\maketitle
\begin{abstract}
Code generation from text requires understanding the user's intent from a natural language description and generating an executable code snippet that satisfies this intent. While recent pretrained language models demonstrate remarkable performance for this task, these models fail when the given natural language description is under-specified. In this work, we introduce a novel and more realistic setup for this task. We hypothesize that the under-specification of a natural language description can be resolved by asking clarification questions. Therefore, we collect and introduce a new dataset named \textbf{CodeClarQA} containing pairs of natural language descriptions and code with created synthetic clarification questions and answers. The empirical results of our evaluation of pretrained language model performance on code generation show that clarifications result in more precisely generated code, as shown by the substantial improvement of model performance in all evaluation metrics. Alongside this, our task and dataset introduce new challenges to the community, including when and what clarification questions should be asked. Our code and dataset are available on GitHub.\footnote{\url{https://github.com/UKPLab/codeclarqa}}
\end{abstract}

\input{tex/1-intro}
\input{tex/2-dataset}
\input{tex/3-pipeline}
\input{tex/4-experiments}
\input{tex/5-results.tex}
\input{tex/6-analysis.tex}
\input{tex/10-related_work}
\input{tex/7-conclusions}
\input{tex/8-limitation}
\input{tex/9-ethical_concern.tex}
\input{tex/11-acknowledgement.tex}

\bibliography{anthology,custom}
\bibliographystyle{acl_natbib}

\input{tex/appendix}

\end{document}

%% file: tex/1-intro.tex
\section{Introduction}
Text-to-code generation aims to understand a user's intention represented by a natural language description (NLD) to generate a code that satisfies the user's intention. 
Models for this task are a crucial component of digital pair-programmers, which assist data scientists \citep{agashe-etal-2019-juice,liu-etal-2021-haconvgnn-hierarchical}, software developers \citep{Chen2021EvaluatingLL,xu2022systematic}, and computer programming educators \citep{Li2022CompetitionLevelCG}.

Recent work addresses this task using pretrained language models (PLMs) fine-tuned on large-scale code data in general-purpose programming languages, such as Python and Java \citep{Chen2021EvaluatingLL,Li2022CompetitionLevelCG,Nijkamp2022ACP,chowdhery2022palm,xu2022systematic,Lahiri2022InteractiveCG}. 
\begin{figure}[!t]
    \centering
    \subfloat[ ]
    {\label{subfig:sub1}
     \includegraphics[width=7.5cm]{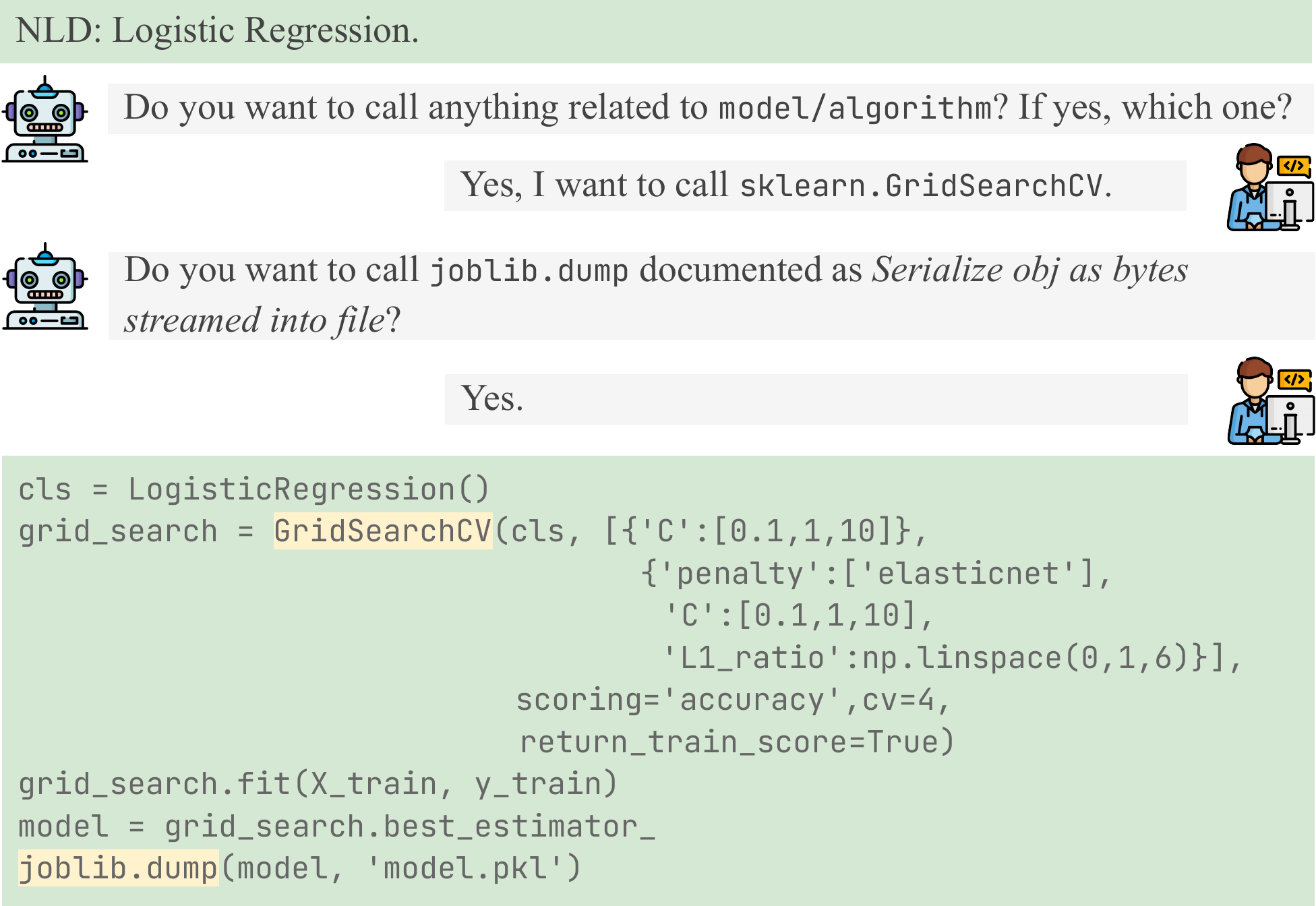}} 
     
    \subfloat[ ]
    {\includegraphics[width=7.5cm]{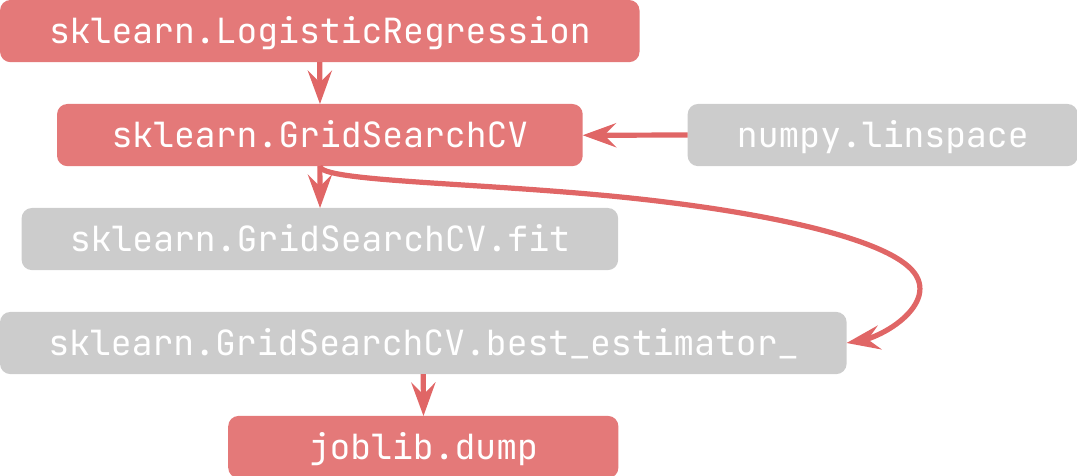}
     \label{subfig:sub2}}
\caption{(a) An example of NLD-code pair that requires further clarification. We highlight operations that need clarification. 
(b) The generated graph of the example. Each node is an operation, with key operations marked in red and the rest in gray. 
Edges show the data flow.}
\label{fig:example}
\end{figure}

\begin{figure*}[!t]
\centering
\resizebox{\textwidth}{!}{
\includegraphics{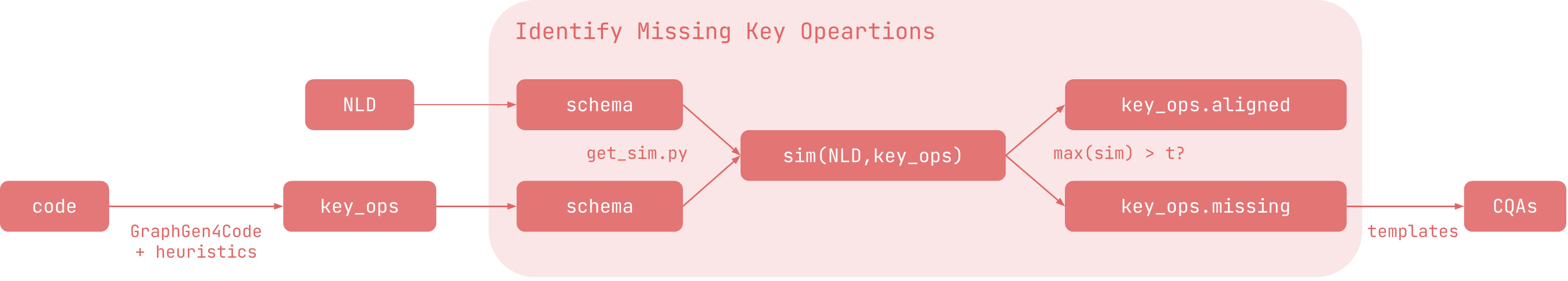}
}
\caption{Our method for creating the CodeClarQA dataset. 
We identify key operations and corresponding documentation from the code. 
We represent them in a latent space using their schemata, letting us compute similarity scores of all schema element pairs between an NLD and the documentation of a key operation. 
If there is an element pair with a similarity score lower than a threshold $t$ the key operation is missing in NLD. 
We adopt templates to create CQAs for missing key operations.}
\label{fig:data_collection}
\end{figure*}

Although these models are successful, they fail to resolve the case of an NLD that lacks enough specifications.
Figure~\ref{subfig:sub1} depicts an example of under-specified NLD (shown in yellow).
The problem of missing specifications in NLDs not only widely occurs in real-world use cases \cite{Lahiri2022InteractiveCG,chaurasia-mooney-2017-dialog} but is also important for training text-to-code generation models. 
Although important, alleviating the under-specification of NLDs is challenging for two reasons. 
First, missing specifications can happen at various levels, including individual operations, argument values of the operations, and sub-tasks consisting of several operations decomposed from the entire source code file as the task. 
Second, it is not obvious how to identify if an NLD carries information about specifications at any level mentioned above or not.

In this paper, we introduce interactivity into text-to-code generation for specifications on the level of individual operation calls in Python.
We hypothesize that by gathering more specifications using these interactions, we can alleviate the incompleteness of NLD's specifications and thus generate a more precise code (Figure~\ref{subfig:sub1}).
To train and evaluate such models, we introduce the CodeClarQA dataset collected through a novel method to synthetically generate  clarification questions and answers (CQAs) for an NLD-Code pair.
To map the operations to natural language, we retrieve the API documentation. 
If there is a low similarity between the NLD and the operation's documentation, we identify the operation as a missing specification and generate a clarification question (CQ). 
The answers to CQs are selected from the given code. 
Furthermore, we propose a pipeline to demonstrate the use case of our dataset for developing NLD-Code generation models. 
Our pipeline consists of three modules -- a clarification need predictor, a CQ generator, and a code generator. 
For each module, we introduce models that can serve as baselines. 

To evaluate the quality of our dataset, we conduct a human evaluation. 
We also evaluate the models we proposed for each component of our pipeline. 
Our empirical results show that by conditioning PLM-based code generation models on CQAs in our dataset, the performance of these models increases, indicating the correctness of our hypothesis and our collected dataset. 
Alongside, our experimental results show that advanced PLMs (e.g., RoBERTa, BART, T5, and CodeT5) struggle to achieve high performance under the interactive code generation pipeline. This important observation demonstrates the difficulty of our dataset for the recent PLMs, introducing new challenges for these models. 

%% file: tex/2-dataset.tex
\section{Creating the CodeClarQA Dataset}
\label{sec:dataset}
We aim to make the process of code generation interactive such that by asking CQs about a given NLD, we resolve NLD under-specification before generating any code. 
To do so, we design a novel method to synthetically collect CQAs for a given NLD-Code pair, leading to the new dataset, which we call CodeClarQA.
Figure~\ref{fig:data_collection} shows a general view of our data creation method. 

\subsection{Identifying Key Operations} 
\label{method_part1}
Key operations correspond to sub-tasks decomposed from the code snippet as the task.
For instance, the code in Figure~\ref{fig:example} can be decomposed into three sub-tasks: call the logistic model, use grid search to fit different logistic models, and save the best model. The corresponding key operations are \textit{sklearn.LogisticRegression}, \textit{sklearn.GridSearchCV}, and \textit{joblib.dump}.
Ideally, an NLD should provide enough information about key operations in a code. 
If the NLD does not do so, it lacks sufficient specifications. 
Thus, our first step for generating CQAs for a given NLD-code pair is to identify the key operations required to generate the code from NLD.

To identify key operations, we represent the code as a graph.  
Data flow graphs are an effective structural representation of code semantics in a code. Given this fact, we use the graph defined by the GraphGen4Code toolkit ~\cite{abdelaziz2021graph4code}, the state-of-the-art toolkit for generating a graph from a source code, including API-related operations and the data flow. This makes it easy for us to identify key operations.
Figure~\ref{subfig:sub1} shows the graph representation. 
Non-leaf nodes represent key operations.
Edges indicate the data flow of the operations. 
For each NLD-Code pair, we parse the code to generate a graph.

Given a graph, we identify nodes with the following properties as key operations:
\textbf{(i) For operations that are one object/function and its methods/fields, we treat the object/function as a key operation.}
    This is coherent with one hypothesis behind the design of GraphGen4Code, where an object/function is first initiated before fields/methods thereof are applied.
    For instance, \textit{sklearn.GridSearchCV} is the key operation among all operations related to it, as other operations apply a method (\textit{.fit}) or read a field (\textit{.best\_estimator\_}) of it (Figure~\ref{subfig:sub2}).
 \textbf{(ii) For a multiple-operation line of code, the last operation on the data flow path is a key operation.} 
    For instance, \textit{sklearn.GridSearchCV} and \textit{numpy.linspace} are in the same line. \textit{sklearn.GridSearchCV} is a key operation since \textit{sklearn.GridSearchCV} is the line's highest-level operation (Figure~\ref{subfig:sub2}). See Appendix~\ref{sec:appendix_1} for details of the procedure of identifying key operations.

\subsection{Is a Key Operation Missing in NLD?} \label{method_part2}

Given a set of key operations required to generate a code, we should identify if the given NLD provides any information about these operations. 
To do so, for each key operation, we propose to align the schema of textual documentation of a key operation with the schema of a given NLD.
A schema \cite{majumder-etal-2021-ask} is defined as a set of important elements of a document. 
Every schema element is either in the form of
(\emph{verb}, \emph{key\mbox{-}phrase}, \emph{relation})
 or 
(\emph{key\mbox{-}phrase}), where \emph{key\mbox{-}phrase} is extracted using YAKE \citep{campos-etal-2020-yake}, and \emph{verb} and \emph{relation} are obtained by searching through the closest verb and its dependency relation using the dependency tree \citep{qi-etal-2020-stanza}. 
An example of (\emph{verb}, \emph{key\mbox{-}phrase}, \emph{relation}) is (transforms, final estimator, \textit{obl}), and an example of (\emph{key\mbox{-}phrase}) is  (pipeline).

For each key operation required to generate a code, we compute similarity scores for all schema element tuples using elements from the NLD and the documentation. 
For each pair of schema elements, we use a pretrained text encoder \citep{reimers-gurevych-2019-sentence} to compute similarity scores between these phrases as key information.
Note that we combine \emph{verb} and \emph{key\mbox{-}phrase} if the schema element is in the triplet form before computing the similarity score.
Eventually, we identify a key operation is missing in the NLD if the highest similarity score of all schema element pairs is lower than the threshold $t$. 
Each key operation is then labeled as \textit{aligned} or \textit{missing}. 
We perform a grid search to find the best $t$ on a validation set, maximizing the F1 score. See Appendix~\ref{sec:appendix_2} for an example.

\subsection{Generating CQAs for Missing Key Operations}
\label{method_part3}
We formulate CQs as multiple-choice questions and yes/no questions. 
The former needs an answer with yes/no following a choice of an API call. 
The latter requires only an answer with yes/no.
\paragraph{Multiple-choice.}
We collect all extracted key operations from the dataset, mentioned or missing, that contain 1023 different API sub-modules, methods, and fields. 
We then extract the last tokens from each operation name, filter out all stop words, and keep operations that share the same last token of their names.
For instance, \textit{sklearn.partial\_fit} and \textit{sklearn.fit} share the same last token as \textit{fit}. 
Note that we hypothesize that for operations with the same name but from a different library, e.g., \textit{keras.fit} and \textit{sklearn.fit}, they refer to the same operation.
We generate multiple-choice questions for these key operations if they are missing.
To do so, we use the template \textit{Do you want to call anything related to \textsc{last\_token}? If yes, which one?}.

\paragraph{Yes/No.}
For operations that do not belong to multiple-choice questions, we generate a yes/no question using the template \textit{Do you want to call \textsc{operation\_name} documented as \textsc{doc}?}. 
For instance, a CQ about \textit{numpy.logspace} is generated as ``Do you want to call \textit{numpy.logspace} documented as \textit{Return numbers spaced evenly on a log scale?}''

\subsection{Dataset}
We use NLD-Code pairs in the notebookCDG dataset \citep{liu-etal-2021-haconvgnn-hierarchical} to create CQAs because of the high code quality ensured by votes of the Jupyter Notebooks and the high NLD quality ensured by the preprocessing method based on the study of markdown documentation \citep{10.1145/3411763.3451617}.
We first identify key operations (\S\ref{method_part1}) and label them as either \textit{aligned} or \textit{missing} (\S\ref{method_part2}). Finally, we select NLD-Code pairs with at most five missing key operations, duplicate missing key operations, and create CQAs (\S\ref{method_part3}).
 \begin{table}[!t]
\small
\begin{center}
\begin{tabular}{@{}lrrrr@{}}
\toprule
 & Total & Train & Dev & Test \\ 
\midrule
\# NLD-Code Samples & 19368 & 17431 & 968 & 969 \\
Avg. NLD Length & 12.43 & 12.45 & 12.22 & 12.30 \\
Avg. Code Length & 44.40 & 44.52 & 44.52 & 42.25 \\
\# Samples w/ CQAs & 12339 & 11098 & 630 & 611 \\
\# Samples w/o CQAs & 7029 & 6333 & 338 & 358 \\
\# CQAs & 17506 & 15711 & 923 & 872 \\
\# Multiple-Choice Qs & 8952 & 8008 & 474 & 470\\
\# Yes/No Qs & 8554 & 7703 & 449 & 402\\
\# Operations & 817 & 749 & 227 & 215 \\
\# Packages & 89 & 82 & 33 & 28 \\
\bottomrule
\end{tabular} 
\end{center}
\caption{Some statistics of our dataset.}
\label{tab:dataset_stat}
\end{table}
Table \ref{tab:dataset_stat} shows dataset statistics. 

%% file: tex/3-pipeline.tex
\section{Pipeline for CQ-driven Code Generation}
\label{sec:pipeline}
Our system generates precise code by asking CQs before generating. To do so, it uses an interactive code generation pipeline that includes three modules: 
(i) a clarification need predictor, 
(ii) a CQ ranker, and 
(iii) a code generator. 

Given an NLD, the clarification need predictor predicts the need to ask CQs, with labels \textit{Need} and \textit{No Need}.
If there is a need for asking CQs, the CQ ranker selects $n$ CQs. 
We set $n$ as five to push these models to choose CQs with the most information gains. 
Given the NLD, CQs and corresponding answers, the code generator generates a code.

%% file: tex/4-experiments.tex
\section{Experiments}
\label{sec:experiments}

Having proposed our dataset (\S\ref{sec:dataset}) and a pipeline for interactive code generation (\S\ref{sec:pipeline}), we next evaluate the quality of the dataset creation method by focusing on \S\ref{method_part2} and results of use our dataset to evaluate recent PLM-based models for each pipeline module for interactive code generation, before assessing the quality of the pipeline. The dataset evaluation analyzes the effectiveness of identifying key operations, while experiments on the pipeline aim to validate our hypothesis that interactiveness helps code generation and evaluate task difficulty.

\subsection{Dataset Evaluation}
To evaluate our dataset creation method, we randomly split our dataset into train/validation/test sets. 
We asked two Ph.D. students in computer science to annotate each NLD-Code pair in the validation and test sets.
The annotation for each NLD-Code pair is a binary label, indicating if the NLD misses any key operation from the code. 
These annotations let us (i) study the properties of our dataset and 
(ii) evaluate the quality of our method for finding missing key operations using different text encoders. See Appendix \S\ref{sec:appendix_4} for more details.

\paragraph{Setting.}
The validation and test set consist of 100 NLD-Code pairs respectively.
The Fleiss Kappa is 0.74 (0.83 for the validation and 0.66 for the test set). We randomly chose one annotator's annotation as reference labels. See Appendix \S\ref{paragraph:discrepancy} for more analysis on annotation results.

\subsection{Clarification Need Predictor}
\label{part:pred_clar_need} 

In order to label when CQs were needed, we learned a binary classifier. This classifier predicts, for an NLD, whether it needs further clarification. The classifier was trained on the NLD-Code pairs in the training portion of the \textbf{CodeClarQA} dataset.


\paragraph{Setting.}
We fine-tune baseline pretrained transformer classifiers, including BERT \citep{devlin-etal-2019-bert}, RoBERTa \citep{liu-etal-2019-roberta}, and the encoder of BART \citep{lewis-etal-2020-bart}. To include models trained on NLD data, we also fine-tune the encoder of PLBART \citep{ahmad-etal-2021-unified}.  Models are fine-tuned on the training set with NLDs as the input. We fine-tune each model for 10 epochs with learning rate $5\times 10^{-5}$ and pick the best-performing model on accuracy. 
We compare the models on the test set using accuracy, precision, recall, and F1.

\subsection{CQ Ranker}
Given an NLD, a CQ ranker should recommend potential key operations by asking CQs. We formulate this as a ranking task, where we select a subset of CQs from a universal set of CQs.
We use all created CQs using our method mentioned in \S\ref{sec:dataset} as the universal set. 

\paragraph{Setting.}
We follow \newcite{aliannejadi-etal-2021-building} 
and fine-tune cross-encoders on all NLD-CQ pairs and experiment with models used in \S\ref{part:pred_clar_need}. Given an NLD-CQ pair, each model is trained to do binary classification. At inference time, all CQs in the universal set are paired with a given NLD and ranked by model score. Given an NLD, positive samples CQs created in the dataset. 
To create negative samples, we experiment with random negative sampling and BM25 \citep{robertson1995okapi}. 
The number of negative samples selected is the average number of positive samples. Each model is trained for 10 epochs with learning rate $5 \times 10^{-5}$. We evaluate model performance with the test set on  $R@k, k\in \{1,3,5,10\}$.

\begin{table}[!t]
\small
\begin{center}
\begin{tabular}{@{}lcccc@{}}
\toprule
Model &  Acc & P & R & F1 \\ 
\midrule
$\text{SentenceT5}_\text{large} (0.83)$  & 87.40 & 71.43 & 80.65 & 75.76 \\
$\text{SentenceT5}_\text{xl} ( 0.82)$  & 87.40 & 68.57 & 82.76 & 75.00 \\
$\text{GTR}_\text{large} (0.70)$   & 88.19 & 71.43 & 83.33 & 76.92 \\
$\text{GTR}_\text{xl} (0.69) $  & 88.98 & 68.57 & \bf{88.89} & 77.42 \\
$\text{MiniLM}_\text{L12} \text{-all-v1} (0.47)$    & 87.40 & 74.29 & 78.79 & 76.47 \\
$\text{MiniLM}_\text{L12} \text{-all-v2} (0.49)$  & 87.40 & 73.53 & 78.12 & 77.61\\ 
$\text{DistilRoBERTa} (0.49) $   & 88.19 & 74.29 & 83.33 & 76.92\\
$\text{RoBERTa}_\text{large} \text{-all} (0.46)$   & 89.76 & 80.00 & 82.35 & 81.16 \\
$\text{MPNet}_\text{base} \text{-all-v1} (0.40)$   & 84.25 & 80.00 & 68.29 & 73.68 \\
$\text{MPNet}_\text{base} \text{-all-v2} (0.42)$  & 86.61 & \bf{82.86} & 72.5 & 77.33 \\
$\text{MPNet}_\text{base} \text{qa-dot} (0.62) $   & 89.76 & \bf{82.86} & 80.56 & 81.69 \\
$\text{MPNet}_\text{base} \text{qa-cos} (0.41)$   & \bf{90.55} & \bf{82.86} & 82.86 & \bf{82.86} \\
\bottomrule
\end{tabular} 
\end{center}
\caption{
Results for identifying missing key operations on our test set using different text encoders. 
The numbers in parenthesis refer to the threshold  optimized on the human-annotated validation set for F1 score.
}
\label{tab:identifying_missing_specifications_results}
\end{table}

\subsection{Code Generator}
The key hypothesis of our work is that interactive code generation systems outperform non-interactive ones. 
In this experiment, we conduct a proof-of-concept experiment to validate this hypothesis, assuming a perfect interactive system with perfectly asked CQs and answers. 
We fine-tune models with and without oracle CQAs from our dataset.
Note that for both yes/no and multiple-choice questions, we have only positive answers in our dataset. 

\paragraph{Setting.}
We experiment with models mentioned by \newcite{Zhou2022DocCoderGC} for fine-tuning, including
GPT-Neo-\{125M, 1.3B\} \citep{gpt-neo}, T5 \citep{DBLP:journals/corr/abs-1910-10683}, and CodeT5 \citep{wang-etal-2021-codet5}. We include CodeParrot-\{110M,1.5B\} \citep{tunstall2022natural}. 
Note that for CodeParrot-110M, we use the model fine-tuned on text-to-code generation.\footnote{\url{https://huggingface.co/datasets/codeparrot/github-jupyter-parsed}}  
Moreover, we finetune PLBART-base \citep{ahmad-etal-2021-unified}. 
We train each model for 40 epochs with learning rate $5\times 10^{-5}$. Each experiment takes up to 6 hours on a single A100 GPU.
We evaluate models on BLEU score \citep{papineni-etal-2002-bleu}, CodeBLEU score \citep{https://doi.org/10.48550/arxiv.2009.10297}, and Exact Match (EM). Note that we don't include state-of-the-art execution-based metrics \citep{huang-etal-2022-execution,https://doi.org/10.48550/arxiv.2212.09248}, since it requires us to include code context into the dataset, which leverages the difficulty of dataset construction. As we don't include code context into the dataset, code predictions are more likely to fail on e.g. variable naming, which affects the execution results but does not necessarily lead to poor code quality.

\subsection{End-to-end Pipeline Evaluation}\label{sec:pipeline_evaluation}
To assess the performance of the entire pipeline (\S\ref{sec:pipeline}), we use the best-performing models for each module.
We pass an NLD to the clarification need predictor. 
Given a positive prediction, we pass the NLD to the CQ ranker.  
For each NLD, we select the top-$k$ ($ k \in \left\{ 1,3,5 \right\}$) ranked CQs by the CQ ranker. 
We compare them to CQs created using our approach and select overlapping CQs.
Finally, we concatenate the NLD and all selected CQs with corresponding answers and feed them to the code generator. 

%% file: tex/5-results.tex
\section{Results}
\label{sec:results}

\subsection{Dataset Evaluation}
We first evaluate the effect of different text encoders on the performance of our method for identifying missing operations. 
Table~\ref{tab:identifying_missing_specifications_results} shows the results. 
We achieve the best performance using $\text{MPNet}_\text{base} \text{qa-cos}$ text encoder. 
We then use our annotations to analyze the predictions of this model. 
Table~\ref{tab:error_analysis} shows the results of this analysis in terms of False Positive (FP) and False Negative (FN) errors.
\begin{table}[!t]
\small
\begin{center}
\begin{tabular}{@{}lcccc@{}}
\toprule
Error & \multicolumn{2}{c}{Freq} & \multicolumn{2}{c}{ER (\%)}\\
 & Dev & Test & Dev & Test \\ 
\midrule
Taxonomy (FP) & 3 (.33) & 3 (.50) & 7.32 & 8.57 \\
Element Pair (FP) & 3 (.33) & 3 (.50) & 7.32 & 8.57 \\
Argument (FN) & 4 (.57)& 4 (.67) & 4.08 & 4.35 \\
\bottomrule
\end{tabular} 
\end{center}
\caption{Statistics of the most common FP and FN predictions. 
Freq refers to frequency, with relative frequency included in the parenthesis. Error rates (ER) are computed on the corresponding predictions. 
}
\label{tab:error_analysis}
\end{table}
For the sake of brevity, we report the full list in Appendix \S\ref{sec:appendix_4}.

\begin{table*}[!t]
\small
\addtolength{\tabcolsep}{-3pt}
\begin{center}
\begin{tabular}{L{0.8cm} L{2.0cm} L{12.2cm}}
\toprule
\textbf{Type} & \textbf{Category} & \textbf{Example}\\

\midrule[.04em]

FP
&
Taxonomy
&
\textbf{NLD}: We've addressed a lot of the issues holding us back when using a \greenbox{linear model}...
\par
\textbf{Code Line:} LCV = \redbox{LassoCV()}
\par
\textbf{Doc:} Lasso CV: Lasso \greenbox{linear model} with iterative fitting along a regularization path. 
\\

\midrule[.02em]
FP 
&
Element Pair
&
\textbf{NLD}: ...we concatenate the two sets while remembering the \greenbox{index} so we can split it later again.
\par
\textbf{Code Line:} train\_features = train\redbox{.drop(['SalePrice'], axis=1)}
\par
\textbf{Doc:} drop: \greenbox{Make} new \greenbox{Index} with passed list of labels deleted.
\\

\midrule[.02em]
FN
&
Argument
&
\textbf{NLD}: Transforming some numerical variables.
\par
\textbf{Code Line:} all\_data['MSSubClass'] = all\_data['MSSubClass']\redbox{.apply(str)}
\par
\textbf{Doc:} apply: Apply a function along an axis of the Data Frame.
\\

\toprule
\end{tabular}
\end{center}
\caption{Examples of predictions in identifying missing key operations. We provide true positive (TP), false positive (FP), and false negative (FN) examples. \textbf{Category} refers to the assigned category of prediction by human evaluation. Key operations and schema element pairs with the highest similarity scores are highlighted.
}
\label{tab:examples_tp_fp_fn}
\end{table*}

\begin{table}[!t]
\small
\begin{center}
\begin{tabular}{@{}lcccc@{}}
\toprule
Model & Acc & Precision & Recall & F1 \\ 
\midrule
$\text{RoBERTa}_\text{base}$  & 64.94 & 65.91 & \bf{95.29} & 77.43 \\
$\text{BART}_\text{base}$ & 70.33 & 74.78 & 79.95 & 77.24 \\
$\text{PLBART}_\text{base}$ & 71.05 & \bf{75.86} & 79.34 & 77.56 \\
$\text{BERT}_\text{base}$ & \bf{71.49} & 75.72 & 80.73 & \bf{78.13} \\
\bottomrule
\end{tabular} 
\end{center}
\caption{Results of clarification need prediction. All numbers are averaged across four runs.}
\label{tab:predict_clarification_need}
\end{table}

The ``Taxonomy'' and ``Element Pair'' error types take up to 7.32\% and 8.57\% of all operations predicted as \textit{aligned} in the validation/test sets, respectively.

The rare case of FP predictions suggests that our approach to generating CQAs effectively creates CQAs for missing key operations. 
The \textit{Taxonomy} error relates the differences related to the taxonomy of terms that could not be identified, taking up to about 8.57\%.
The \textit{Element Pair} error relates to the cases where non-relevant schema elements are aligned, taking up to about 8.57\%.  
The \textit{Argument} error represents the alignment between arguments,  taking up only 4.08\%/4.35\% of all negative predictions from the validation/test set. 
Table~\ref{tab:examples_tp_fp_fn} shows examples of these errors.

For the taxonomy error,  our method identifies a schema element match of \textit{linear models} but fails to predict the difference between a \emph{lasso linear model} and a \emph{linear model} in the taxonomy of machine learning terms. This finding shows a potential direction of future work, in which \textit{aligned} operations might require clarification to be distinguished from operations with similar names.
The example of \textit{Argument} error reflects the case where a complete semantics of the operation needs both the documentation and the argument values.
As we proposed to compare documentation and the NLD, we miss out on arguments that can complement the semantics of the operation. 
The corresponding example shows that the operation \textit{.apply} 's semantics is incomplete without the argument \textit{str}. 
This is reflected in the design of our method, as we use API documentation which reflects the semantics of the API call, while argument values are not documented.

The \textit{Element Pair} error example shows that (make, index, \textit{obj}) from the documentation's schema is aligned with (index) from NLD's schema. 
In contrast, the key operation from the documentation should be either \textit{drop} or \textit{deleted}. 

\subsection{Clarification Need Predictor Evaluation}
Table~\ref{tab:predict_clarification_need} summarizes the results of different classifiers. Most tested models obtain relatively high performances except for $\text{RoBERTa}_\text{base}$, which overfits the imbalanced data where 63.71\% samples have positive labels, as shown by the high recall but low precision. Moreover, $\text{BERT}_\text{base}$ has the best performance on both accuracy and F1 score. 

\begin{table}[!t]
\small
\begin{center}
\begin{tabular}{@{}lccccc@{}}
\toprule
 &   \multicolumn{5}{@{}c@{}}{$R@k$(\%)}\\
Model & NS &  1 & 3 & 5 & 10 \\
\midrule
BM25 & & 0.43 & 0.79 & 0.79 & 1.22 \\
\multicolumn{1}{@{}l}{$\text{RoBERTa}_\text{base}$} 
& \hspace*{1mm}\textit{BM25}  & 0.09 & 0.26 & 0.73 & 1.61 \\
&\hspace*{1mm}\textit{Rand.} & 5.3 & 12.98 & 18.45 & 27.02 \\ 
\multicolumn{1}{@{}l}{$\text{BERT}_\text{base}$ }
&\hspace*{1mm}\textit{BM25} & 5.21 & 13.17 & 17.47 & 24.39 \\
&\hspace*{1mm}\textit{Rand.} & 4.57 & 12.41 & 17.82 & 26.88 \\
\multicolumn{1}{@{}l}{$\text{PLBART}_\text{base}$} 
&\hspace*{1mm}\textit{BM25} & 4.41 & 10.84 & 15.85 & 23.19 \\
& \hspace*{1mm}\textit{Rand.} & 10.64 & 17.91 & 22.24 & 30.15 \\
\multicolumn{1}{@{}l}{$\text{BART}_\text{base}$} 
& \hspace*{1mm}\textit{BM25} & 4.88 & 10.98 & 14.66 & 21.44 \\
&\hspace*{1mm}\textit{Rand} & \bf{13.62} & \bf{21.34} & \bf{26.19} & \bf{34.88} \\ 
\bottomrule
\end{tabular} 
\end{center}
\caption{Results of CQ ranking on the test set. Column headers indicate the $k$ value in $R@k$ in percentages.
\textit{NS} refers to the negative sampling strategy. All numbers are averaged across four runs.
}
\label{tab:ranking}
\end{table}

\subsection{CQ Ranker Evaluation}\label{sec:analysis_ranker}
We report the results of our experiments on CQ generation in Table~\ref{tab:ranking}. The results confirm that our design of selecting CQs is reasonable, with the best-performing model showing similar results to the ``Question Relevance'' task designed by \newcite{aliannejadi-etal-2021-building}. However, we hypothesize that our task is more challenging, as the lexical overlap between the NLD and the correctly selected CQs is low due to our design of dataset creation which looks for key operations with documentation that has no keyword matches to the NLD. This requires the model to utilize the under-specified NLD and infer the topic of the task and the user's intent before providing suggestions by asking CQs.

Our hypothesis is strongly supported by the low recall of the BM25 ranker, which ranks CQs based on their lexical similarities with NLD. Moreover, we find that models trained with the BM25 negative sampler always perform lower than the ones trained with the random sampler, which also supports our hypothesis because the BM25 negative sample is expected not to select CQs that have high lexical overlap with the NLD as negative samples, while they have a higher chance of asking key operations that are ``mentioned''. 

\begin{table}[!t]
\small
\begin{center}
\begin{tabular}{@{}lccc@{}}
\toprule
Method & BLEU & CodeBLEU & EM(\%) \\ 
\midrule
$\text{T5}_\text{base}$& 7.88 & 14.65 & 0.88 \\
$\text{T5}_\text{base}$\textit{+CQAs}& 12.43 & 19.04 & 2.09 \\
$\text{GPT-Neo}_\text{125M}$& 11.89 & 24.75 & 0.00 \\
$\text{GPT-Neo}_\text{125M}$\textit{+CQAs}& 15.63 & 26.97 & 0.00 \\
$\text{GPT-Neo}_\text{1.3B}$& 13.95 & 26.57 & 0.00 \\
$\text{GPT-Neo}_\text{1.3B}$\textit{+CQAs}& 19.64 & 31.05 & 0.00 \\
$\text{CodeParrot}_\text{110M}$& 12.61 & 26.42 & 0.10 \\ 
$\text{CodeParrot}_\text{110M}$\textit{+CQAs}& 17.97 & 31.01 & 0.00\\
$\text{CodeParrot}_\text{1.5B}$& 12.04 & 26.02 & 0.10 \\ 
$\text{CodeParrot}_\text{1.5B}$\textit{+CQAs}& 17.77 & 30.74 & 0.00\\
$\text{PLBART}_\text{base}$ & 24.63 & 28.04 & 12.00 \\
$\text{PLBART}_\text{base}$\textit{+CQAs}& 38.91 & 38.54 & \bf{18.03} \\
$\text{CodeT5}_\text{base}$ & 27.03 & 32.66 & 10.84 \\ 
$\text{CodeT5}_\text{base}$\textit{+CQAs}& \bf{39.13} & \bf{38.99} & 13.93\\
\bottomrule
\end{tabular} 
\end{center}
\caption{Code generation results without and with created CQAs in our dataset. All numbers are
averaged across four runs.
}
\label{tab:generation}
\end{table}

\subsection{Code Generator Evaluation}\label{sec:codegen_results}
We train recent models using only the NLD-Code pairs or with NLD, Code, and  CQAs in the \textbf{CodeClarQA} dataset. The experimental setup aims to test our hypothesis that interactiveness helps code generation by running code generation models with ``perfect'' clarifications. Note that this only serves as proof of concept, as CQAs contain operation names in the target source code, leading to data leakage because the names of the API calls exist in the CQs. 

Table~\ref{tab:generation} shows that all models fine-tuned with CQs have substantially better performance, with the largest gap of 14.28 in BLEU, 10.5 in CodeBLEU, and 6.03 in EM reached by $\text{PLBART}_\text{base}$, which supports our hypothesis that interactions help code generation. Moreover, all models pretrained on code data have better performances, with $\text{CodeT5}_\text{base}$ and $\text{PLBART}_\text{base}$ as the best-performing models we tested.

\subsection{Pipeline Evaluation}\label{sec:pipeline_results}

We use $\text{BERT}_\text{base}$ clarification need predictor, $\text{BART}_\text{base}$ CQ ranker with random negative sampling, and $\text{PLBART}_\text{base}$ trained with \textit{CQAs}. Given the question ranker's predictions, we select CQAs from the test sample with CQ included in the 
top-$k$ ($ k \in \left\{ 1,3,5 \right\}$) list yielded by the CQ ranker.
Besides concatenating selected CQs to NLDs, we also concatenate CQs without selecting them, treating them as ``unanswered clarifications''.
\begin{table}[!t]
\small
\begin{center}
\begin{tabular}{@{}m{1.3cm}m{0.05cm}m{0.55cm}m{0.55cm}m{0.55cm}m{0.55cm}m{0.55cm}m{0.55cm}@{}}
\toprule
Model & $k$ & \multicolumn{2}{c}{BLEU} & \multicolumn{2}{c}{CodeBLEU} & \multicolumn{2}{c}{EM(\%)} \\
& & \multicolumn{1}{c}{\xmark} & \multicolumn{1}{c}{\cmark} & \multicolumn{1}{c}{\xmark} & \multicolumn{1}{c}{\cmark} & \multicolumn{1}{c}{\xmark} & \multicolumn{1}{c}{\cmark}  \\ 
\midrule
\multirow{3}{*}{$\text{PLBART}_\text{base}$} 
 & 1 & 19.51 & 19.82 & 22.43 & 22.16 & 5.24 & 6.94 \\
 & 3 & 15.57 & 21.33 & 22.69 & 23.51 & 4.15 & 7.53 \\ 
 & 5 & 13.20 & \bf{22.07} & 21.77 & \bf{24.07} & 3.95 & \bf{7.92} \\
 \midrule
 \multirow{3}{*}{$\text{CodeT5}_\text{base}$}
 & 1 & 19.14 & 24.04 & 24.14 & 25.56 & 5.37 & 7.15 \\
 & 3 & 14.58 & 25.45 & 25.28 & 26.63 & 4.36 & 7.51 \\
 & 5 & 13.03 & \bf{26.27} & 25.13 & \bf{27.24} & 4.33 & \bf{7.82} \\
\bottomrule
\end{tabular} 
\end{center}
\caption{
Pipeline evaluation. \xmark refers to experiments with top $k$ CQs directly appended to the NLD, and \cmark refers to experiments with CQs selected as noted in \S\ref{sec:pipeline_evaluation}. All numbers are
averaged across four runs. 
}
\label{tab:inference}
\end{table}

We report the results of pipeline evaluation in Table~\ref{tab:inference}. We find that model performances on all evaluation metrics substantially increased with more highly-ranked CQs being included and ``answered'' by comparing highly-ranked CQs and the CQAs in the dataset. Moreover, we also find the opposite trend for ``un-answered clarifications'' where models perform worse with more highly-ranked CQs included (but not ``answered''). This aligns with the challenge of asking CQs mentioned in \S\ref{sec:analysis_ranker}.

Last but not least, we compare the pipeline inference results in Table~\ref{tab:inference} to the results in Table~\ref{tab:generation}. Notably, our pipeline underperforms models trained on data with only NLDs and code. This is expected, as we use code generators that are fine-tuned on all CQAs, and the results of ranking CQs suggest that the task of asking CQs is challenging (\S\ref{sec:analysis_ranker}). 

%% file: tex/6-analysis.tex
\section{Analysis}

Intuitively, asking CQs helps code generation because it provides more specifications, thus aligning model generations to desired and better-quality outputs. To test if this hypothesis stands under the context of our proposed task and pipeline, we analyze model generations quantitatively and qualitatively.

\begin{table}[!t]
\small
\begin{center}
\begin{tabular}{@{}lccccc@{}}
\toprule
 & &\multicolumn{4}{c}{$\text{Recall}$}\\  
Model & &\multicolumn{2}{c}{$\text{micro}$} & \multicolumn{2}{c}{$\text{macro}$}\\
\midrule
& & \xmark & \cmark & \xmark & \cmark \\  
\midrule
\multirow{5}{*}{$\text{PLBART}_\text{base}$}
& & \multicolumn{2}{c}{31.08} & \multicolumn{2}{c}{32.89}\\
& \textit{+top 1}
& 14.39 & 25.14 & 15.50 & 23.69\\
& \textit{+top 3}
& 18.69 & 30.79 & 19.33 & 29.00\\
& \textit{+top 5}
& 17.31 & 32.65 & 18.20 & 30.79\\
& \textit{+CQAs}
& \multicolumn{2}{c}{\bf{92.72}} & \multicolumn{2}{c}{\bf{92.23}}\\ 
\midrule
\multirow{5}{*}{$\text{CodeT5}_\text{base}$}
& & \multicolumn{2}{c}{37.44} & \multicolumn{2}{c}{39.17}\\
& \textit{+top 1}
& 15.45 & 28.27 & 17.07 & 27.51\\
& \textit{+top 3}
& 17.72 & 33.62  & 18.90 & 32.47\\
& \textit{+top 5}
& 17.32 & 35.67 & 18.60 & 34.39\\
& 
\textit{+CQAs}
& \multicolumn{2}{c}{\bf{92.71}} & \multicolumn{2}{c}{\bf{92.63}}\\ 
\bottomrule
\end{tabular} 
\end{center}
\caption{Micro and macro recalls of identified missing key operations. \xmark refers to experiments with top $k$ CQs directly appended to the NLD, and \cmark refers 
to experiments with CQs selected as noted in \S\ref{sec:pipeline_evaluation}. All numbers are
averaged across four runs. 
}
\label{tab:key_op_recall}
\end{table}

\paragraph{Recall of identified missing key operations.} 
Table~\ref{tab:key_op_recall} shows the recall of missing key operations from predictions. We find that training with clarifications includes substantially more missing key operations, while the pipeline still does not outperform models trained on data with only NLDs and code, similar to Table~\ref{tab:inference}. Furthermore, we report Pearson correlation between the recall of missing key operations and code generation results (See Table~\ref{tab:key_op_pearson}), finding high and positive correlations which support our hypothesis that asking CQs helps code generation through clarified key operations.

\paragraph{Case study.} We examine predictions and provide an example in Table~\ref{tab:example_generation}. We find that training with oracle CQAs leads to predictions close to the ground truth, especially on operations, with only differences at argument-level specifications, which is expected as we focus on clarifications on operations. However, the task is challenging as the top 5 ranked CQs do not include CQs in the reference CQAs, leading to the pipeline prediction including a call of confusion matrix but missing \textit{AdaBoostClassifier} and \textit{cross\_val\_predict}.

\begin{table}[!t]
\small
\begin{center}
\begin{tabular}{@{}lcccc@{}}
\toprule
 & & \multicolumn{3}{c}{$\rho$}\\
Model & Recall & BLEU & CodeBLEU & EM(\%) \\
\midrule

\multirow{2}{*}{$\text{PLBART}_\text{base}$}
& micro & 0.929$^{*}$ & 0.949$^{*}$ & 0.915$^{*}$\\
& macro & 0.932$^{*}$ & 0.962$^{*}$ & 0.923$^{*}$\\
\midrule
\multirow{2}{*}{$\text{CodeT5}_\text{base}$}
& micro & 0.918$^{*}$ & 0.938$^{*}$ & 0.910$^{*}$\\
& macro & 0.909$^{*}$ & 0.949$^{*}$ & 0.911$^{*}$\\
\bottomrule
\end{tabular} 
\end{center}
\caption{Pearson correlation coefficient ($\rho$) between recalls and results from Table~\ref{tab:generation} and Table~\ref{tab:inference}. Results marked with $^*$ are statistically significant (\textit{p} < 0.001). 
}
\label{tab:key_op_pearson}
\end{table}

\begin{table*}[!t]
\small
\addtolength{\tabcolsep}{-3pt}
\begin{center}
\begin{tabular}{L{2.3cm} L{12.7cm}}
\toprule 
\multicolumn{2}{l}{\textbf{NLD}: Confusion Matrix for the Best Model.}
\\
\midrule[.04em]
Reference CQAs 
& \textbf{CQ1}: Do you want to call anything related to 'model/algorithm'? If yes, which one? 
\par
\textbf{A1}: "Yes, I want to call \redbox{'sklearn.AdaBoostClassifier'}
\par
\textbf{CQ2}: Do you want to call anything related to 'predict'? If yes, which one?
\par
\textbf{A2}: Yes, I want to call \redbox{'sklearn.cross\_val\_predict'}
\\

\midrule[.04em]
Ground Truth &
ada = \redbox{AdaBoostClassifier}(n\_estimators=200, random\_state=0, learning\_rate=0.05)
\par
result = \redbox{cross\_val\_predict}(ada, X, Y, cv=10)
\par
sns.heatmap(confusion\_matrix(Y, result), cmap='winter', annot=True, fmt='2.0f')
\par
plt.show()
\\

\midrule[.02em]
$\text{CodeT5}_\text{base}$ &
y\_pred = model.predict(X\_test)
\par
y\_pred\_classes = np.argmax(y\_pred, axis=1)
\par
y\_true = np.argmax(y\_test, axis=1)
\par
import scikitplot as skplt
\par
skplt.metrics.plot\_confusion\_matrix(y\_true, y\_pred\_classes, title='Confusion Matrix for Best Model')
\par
plt.show()
\\

\midrule[.02em]
$\text{CodeT5}_\text{base}$\textit{+top 5} &
print(confusion\_matrix(y\_test, gbc.predict(X\_test))[1])
\par
print(classification\_report(y\_test, gbc.predict(X\_test))[1])
\\

\midrule[.02em]
$\text{CodeT5}_\text{base}$\textit{+CQAs} &
ada = \redbox{AdaBoostClassifier}()
\par
result = \redbox{cross\_val\_predict}(ada, X, Y, cv=10)
\par
sns.heatmap(confusion\_matrix(Y, result), cmap='winter', annot=True, fmt='2.0f')
\par
plt.show()
\\

\toprule
\end{tabular}
\end{center}
\caption{Example of predictions $\text{CodeT5}_\text{base}$ without asking CQs, with pipeline predictions, and with oracle CQAs. Missing operations and schema element pairs with the highest similarity scores are highlighted. Note that \textit{top 5} ranked CQs do not include CQs in reference CQAs.
}
\label{tab:example_generation}
\end{table*}

%% file: tex/10-related_work.tex
\section{Related Work}

\paragraph{CQ generation.}
\newcite{aliannejadi2019asking,aliannejadi-etal-2021-building} define CQs based on facets/aspects of the text input's topic, guiding annotators to write CQs based on the facet information. 
\newcite{Eberhart2022GeneratingCQ} ask CQs for query refinement based on facets/aspects from existing NLDs in a dataset. Our work is distinguished from the above works as our method does not require a predefined collection of facets/aspects of the text inputs.  
The advantage of our method is that we collect NLDs as specifications from code.

More generally, two main focuses of work on CQ generation are (i) disambiguation of terms \citep{xu-etal-2019-asking,guo2021abgcoqa} and (ii) providing more information \citep{rao-daume-iii-2018-learning,guo2021abgcoqa,majumder-etal-2021-ask,nakano-etal-2022-pseudo}. With the goal of disambiguation of terms, \newcite{xu-etal-2019-asking} utilize the knowledge base to create CQs that disambiguate different entities that share the same entity names. \newcite{guo2021abgcoqa} included CQs of coreference resolution that disambiguate pronouns.
\newcite{rao-daume-iii-2018-learning,guo2021abgcoqa} define CQs to gather information missing from textual input. 
\newcite{majumder-etal-2021-ask} ask CQs on missing information from the item description but existing in similar items, defined as missing schema. \newcite{nakano-etal-2022-pseudo} construct pseudo-CQs by eliminating a part of a sentence and transforming it into a CQ and a corresponding answer. 
Our work adopts the definition of CQs as asking for new information and is distinguished from these works by defining a new type of information as key operations for a code, which are challenging to be defined and identified if they are included in the original text query.

\paragraph{Text-to-Code generation.}
Text-to-code generation was first defined through learning on the parallel corpus of NLD-Code pairs \citep{pmlr-v37-allamanis15,miceli-barone-sennrich-2017-parallel,yin2018mining}. 
To study programming in practice with dependency between different code snippets, \citet{iyer-etal-2018-mapping} introduced a more challenging task that studies generation based on NLD and programming context. \citet{agashe-etal-2019-juice} address the task of generating code cells on Jupyter Notebook given previous markdown and code cells. 
Our work also sources NL-Code pairs collected from Jupyter Notebooks \citep{liu-etal-2021-haconvgnn-hierarchical}. We do not consider dependency between different code/markdown cells when creating CQA, because including previous cells will change the necessity of asking some CQs and make our CQA creation method less controllable. 

Recent research also focuses on generating code utilizing API knowledge or existing source code. 
\citet{xu-etal-2020-incorporating} augment data with samples created by documentation. 
\citet{parvez-etal-2021-retrieval-augmented} retrieve samples from the training set or an archived code database. \citet{Zhou2022DocCoderGC} use retrieval-augmented generation approach by retrieving documentation from source code API usage. 
In contrast, we design the task of retrieving CQs and consider interactivity between the model and the user.

%% file: tex/7-conclusions.tex
\section{Conclusion and Future Work}
In this paper, we introduced a new challenge of asking clarification questions for code generation for Python, along with a method to generate a dataset to create clarification questions and answers that do not require human annotations over the whole dataset. 
We release our collected dataset CodeClarQA, which consists of clarification questions and answers on API usage. 
We further proposed a pipeline system implemented by recent text and code encoders to evaluate model performances on this challenge. 
Our experimental results confirm that clarification questions and answers are strong information-gathering methods for better generation of code while deciding when to ask clarification questions and what questions to ask remains challenging.
Future works include improving clarification questions for higher user engagement and question diversity; studying the lack of user intent completeness beyond the level of operations, e.g., lack of user intent completeness in arguments; and introducing conversational relations between clarification questions.

%% file: tex/8-limitation.tex
\section*{Limitations}

Our method primarily focuses on operation-level specifications, while there are real-world use cases with other specifications.
Moreover, our method of creating CQAs can only be scaled to all Python codes that involve heavy API usage. 
However, if a similar code knowledge graph generator of another language is developed, our method can also be scaled to the corresponding language.
Our method is also limited in identifying specifications missing from the NLD, suggesting potential future work to create CQs about specifications ``mentioned but not specified enough'' in the NLD. 

%% file: tex/9-ethical_concern.tex
\section*{Ethical Concerns}
One concern about the data is the issue of copyright. \newcite{liu-etal-2021-haconvgnn-hierarchical} have checked the data policy of all 20 Kaggle competitions, in which none has copyright issues. Furthermore, they have contacted Kaggle's administrator and have made sure that the dataset collection procedure did not violate the platform's policy. We also check the license of open-source APIs when collecting documentation and make sure that there is no concern about copyright issues. Another concern about the data is that it might include privacy data. Again, we think that our data has a minimum risk of leakage of data with privacy concerns since we only collect data from the 20 Kaggle competitions where there is no concern of privacy data. The API documentation also has the minimum risk of containing data with privacy concerns.

%% file: tex/11-acknowledgement.tex
\section*{Acknowledgements}

We thank Xuye Liu and Dakuo Wang for providing the original dataset and source code for dataset preprocessing. We thank Nico Daheim, Ben Peters, Mert Tiftikci, Kexin Wang, Imbesat Hassan Rizvi, Dominic Petrak for their valuable feedback and suggestions on a draft of this paper. We thank the anonymous reviewers for their detailed and insightful comments. 

This work has been funded by the LOEWE Distinguished Chair “Ubiquitous Knowledge Processing” (LOEWE initiative, Hesse, Germany), by 
EU's Horizon Europe Research and Innovation Actions (UTTER, contract 101070631), and 
by the Funda\c{c}\~ao para a Ci\^encia e Tecnologia through contract  UIDB/50008/2020. 

%% file: tex/appendix.tex
\newpage
\section*{Appendix}\label{sec:appendix}

\appendix

\section{Procedure of Identifying Key Operation.}\label{sec:appendix_1}
We present our procedure for identifying key operations in Algorithm~\ref{pseudocode_ko} as a detailed description of \S\ref{method_part1}. Given an NLD-Code pair and all source codes from its corresponding notebook, our method first extracts operations for the entire notebook and selects operations corresponding to the code from the NLD-Code pair. We then identify key operations by keeping (i) operations from the same API sub-module that have the shortest data flow path and (ii) operations that correspond to the last operation within the same line. Note that we also filter out operations that (i) are print functions, (ii) are numerical operations, and (iii) have no corresponding documentation.

\begin{algorithm*}[!ht]\small
\begin{algorithmic}[1]
    \Function{Extract Key Operations and Documents}{code\_sample, code\_notebook}
        
        \Comment{\textbf{Step 1}: Parse the code to build the graph and get operations for the notebook.}
        
        \State {ops\_notebook $\gets$ GraphGen4Code(code\_notebook).}
        
        \Comment{\textbf{Step 2}: Find all operations corresponding to code from the NLD-Code pair}
        \State {ops\_sample $\gets$ get\_ops\_sample(code\_sample, code\_notebook, ops\_notebook)} 
        
        \Comment{\textbf{Step 3}: Pop operations of either import functions or numerical operations}
        \For{op $\in$ ops\_sample}
            \If {type(ops\_sample) $\in$ \{import\_function, numerical\_expression\}}
                \State ops\_sample.pop(op)
            \EndIf
        \EndFor
        
        \Comment{\textbf{Step 4}: Keep operations that are the last call of the same line of code}
        \For{op $\in$ ops\_sample}
            \If {op is not the last operation of the line}
                \State ops\_sample.pop(op)
            \EndIf
        \EndFor
        
        \Comment{\textbf{Step 5}: Pop operations of print functions}
        \For{op $\in$ ops\_sample}
            \If {type(ops\_sample) $\neq$ print\_function}
                \State ops\_sample.pop(op)
            \EndIf
        \EndFor
        
        \Comment{\textbf{Step 6}: Keep only operations that do not have other operations in its data flow path}
        \State{key\_ops $\gets$ []}
        \For{op $\in$ ops\_sample}
            \If {\{n $|$ n $\in$ ops\_sample $\And$ n $\in$ op.path\} = $\emptyset$}
                \State{key\_ops.append(op)}
            \EndIf
        \EndFor
        
        \Comment{\textbf{Step 7}: Extract Documentations and keep only operations with documentation}
        
        documentations $\gets$ []
        \For{op $\in$ key\_ops}
            \If {op has no corresponding documentation}
                \State key\_ops.pop(op)
            \Else{}
                \State op.documentation $\gets$ get\_documentation(op)
            \EndIf
        \EndFor
        
        \State{\Return{key\_ops}}
        
    \EndFunction
\end{algorithmic}
\caption{Procedure of Extracting Key Operations}
\vspace{-0.4em}
\label{pseudocode_ko}
\end{algorithm*}
\section{Preliminary Experiments on Identifying Missing Key Operations}\label{sec:appendix_2}
We also considered code/documentation-trained models for computing similarities preliminarily. We experimented with RFLHU-BERTOverflow \citep{abdelaziz2022blanca}, which is trained on documentation-StackOverflowPosts pairs and performs similarly to the publicly unavailable RFLHU-CodeBERT in \citet{abdelaziz2022blanca}. We obtained 75.59, 57.14, 55.56, and 56.34 in accuracy, precision, recall, and F1. This is substantially lower than all the results from Table~\ref{tab:identifying_missing_specifications_results}.

\section{Example of Identifying if an Key Operation is Missing}\label{sec:appendix_3}
\begin{figure*}[!ht]
\centering
\resizebox{\textwidth}{!}{
\includegraphics{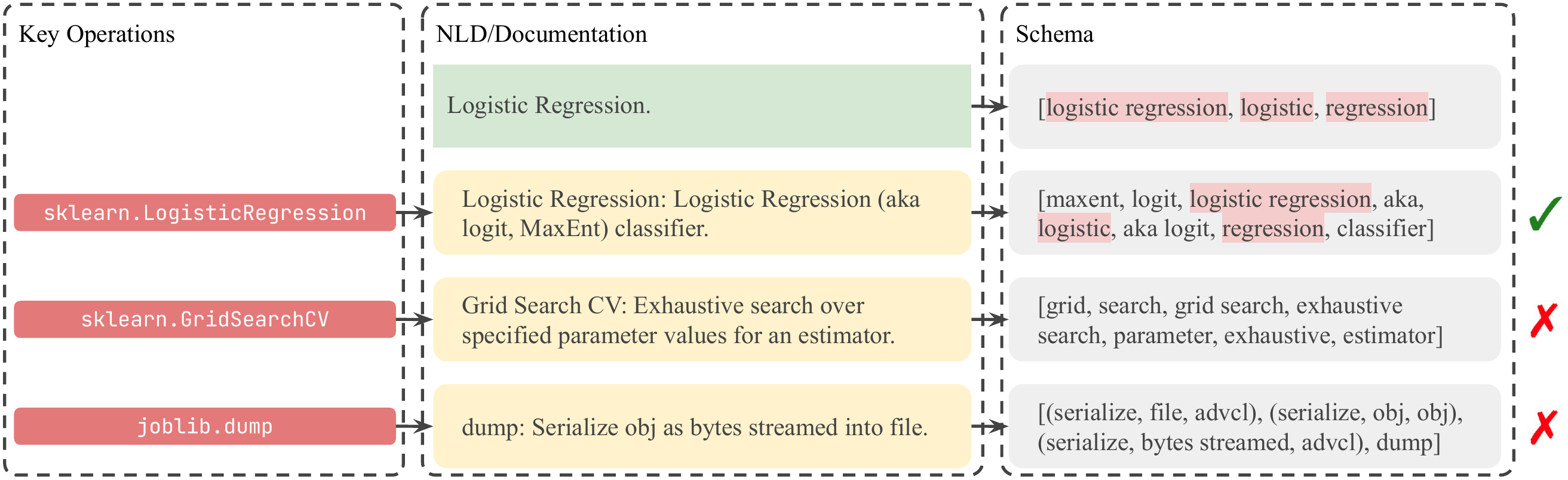}
}
\caption{Example of identifying key operations with the example from Figure~\ref{fig:example}. \cmark means that the key operation is \textit{aligned}, and \xmark means that the key operation is \textit{missing}. Schema element pairs with the highest similarity scores are highlighted if the operation is predicted \textit{aligned}.}
\label{fig:identify_key_op_example}
\end{figure*}

We present an example of identifying if a key operation is missing figure~\ref{fig:identify_key_op_example}. Given the key operations we have extracted (Figure~\ref{subfig:sub2}), we identify if a key operation is missing by comparing all its schema elements with schema elements of the NLD.

\section{Examples of Error Types}\label{sec:appendix_4}
We analyzed predictions of $\text{MPNet}_\text{base} \text{qa-cos}$ text encoder using our annotations. Table~\ref{tab:error_examples_full} shows examples of all types of FP and FN predictions we categorize. We also present in Table~\ref{tab:error_analysis_full} the statistics of all FP and FN predictions. 

\begin{table*}[!ht]
\small
\addtolength{\tabcolsep}{-3pt}
\begin{center}
\begin{tabular}{@{}L{2.5cm} L{8.5cm} L{4cm}@{}}
\toprule
\textbf{Type of Error} & \textbf{Example} & \textbf{Explanation} \\

\midrule[.04em]

Taxonomy (FP)

&
\textbf{NLD}: We've addressed a lot of the issues holding us back when using a \greenbox{linear model}...
\par
\textbf{Code Line:} LCV = \redbox{LassoCV()}
\par
\textbf{Doc:} Lasso CV: Lasso \greenbox{linear model} with iterative fitting along a regularization path. 

& \greenbox{Lasso linear model} should be distinguished from \greenbox{linear model}.
\\

\midrule[.02em]
Element Pair (FP)

&
\textbf{NLD}: ...we concatenate the two sets while remembering the \greenbox{index} so we can split it later again.
\par
\textbf{Code Line:} train\_features = train\redbox{.drop(['SalePrice'], axis=1)}
\par
\textbf{Doc:} drop: \greenbox{Make} new \greenbox{Index} with passed list of labels deleted.

& Method identify \redbox{drop} as non-missing only by seeing \greenbox{index} in both the NLD and the documentation.
\\

\midrule[.02em]

Multiple Operations (FP)

&
\textbf{NLD}: \greenbox{Categorical} Features. Let us look at the missing values in categorical features in detail.
\par
\textbf{Code Line:} categorical\_features.isnull().sum()\redbox{.sort\_values(ascending=False)}
\par
\textbf{Doc:} sort values: Sort the \greenbox{Categorical} by category value returning a new

& Only \redbox{isnull}, \redbox{sum}, \redbox{sort\_values} together refer to  \greenbox{look at the missing values in } \greenbox{categorical features}.
\\

\midrule[.02em]
Model (FP)

&
\textbf{NLD}: The variable importances from the boosted model on the reduced \greenbox{dataset}.
\par
\textbf{Code Line:} sns\redbox{.set\_style('darkgrid')}
\par
\textbf{Doc:} \greenbox{set} style: Set the parameters that control the general style of the \greenbox{plots}.

& Method yields wrong prediction (positive) by comparing \greenbox{dataset} and \greenbox{(set, plots, \textit{obj})}.
\\

\midrule[.02em]
Argument (FN)

&
\textbf{NLD}: Transforming some numerical variables.
\par
\textbf{Code Line:} all\_data['MSSubClass'] = all\_data['MSSubClass']\redbox{.apply(str)}
\par
\textbf{Doc:} apply: Apply a function along an axis of the Data Frame.

& \redbox{apply(str)} corresponds to the NLD, not \redbox{apply} itself.
\\

\midrule[.02em]
Element Missing (FN)

&
\textbf{NLD}: The ' Age', ' Cabin', ' Embarked', ' Fare' columns have missing values.
\par
\textbf{Schema:} [embarked, missing, columns, age, cabin, fare]
\par
\textbf{Code Line:} full['Embarked'].fillna('S', inplace=True)
\par
\textbf{Doc:} fillna: Fill NA/ NaN values using the specified method.
\par
\textbf{Schema:} [(fill, fillna, \textit{nsubj}), (fill, method, \textit{obj})]
& Method fails to extract \greenbox{NA} and \greenbox{NaN} and compare them to \greenbox{missing}.
\\

\midrule[.02em]
Paraphrase (FN)

&
\textbf{NLD}: Train again for all data and submit.
\par
\textbf{Code Line:} rfc\redbox{.fit(X\_train\_all, y\_train\_all)}
\par
\textbf{Documentation:} fit: Fit the calibrated model.
& Model cannot yield high similarity scores between \greenbox{train} and \greenbox{fit}.
\\

\midrule[.02em]
Abbreviation (FN)

&
\textbf{NLD}: GBDT: .
\par
\textbf{Code Line:} gbdt = GradientBoostingClassifier(...)
\par
\textbf{Documentation:} Gradient Boosting Classifier: Gradient Boosting for classification.

& Model cannot yield high similarity scores between \greenbox{gbdt} and \greenbox{Gradient Boosting} \greenbox{Classifier}.
\\

\toprule
\end{tabular}
\end{center}
\caption{Examples of all types of human evaluated errors in the human-annotated validation and test sets. We provide true positive (TP), false positive
(FP), and false negative (FN) examples. Category refers to the assigned category of prediction by human evaluation.
Key operations and schema element pairs with the highest similarity scores are highlighted.
}
\label{tab:error_examples_full}
\end{table*}

\begin{table}[!ht]
\small
\begin{center}
\begin{tabular}{@{}lcccc@{}}
\toprule
Error Type & \multicolumn{2}{c}{Freq} & \multicolumn{2}{c}{ER (\%)}\\
 & Dev & Test & Dev & Test \\ 
\midrule
Taxonomy (FP) & 3 (.33) & 3 (.50) & 7.32 & 8.57 \\
Element Pair (FP) & 3 (.33) & 3 (.50) & 7.32 & 8.57 \\
Multiple Operations (FP) & 2 (.22) & 0 (.00) & 4.87 & 0.00 \\
Model (FP) & 1 (.11) & 0 (.00) & 2.43 & 0.00 \\
\midrule
Argument (FN) & 4 (.57)& 4 (.67) & 4.08 & 4.35 \\
Element Missing (FN) & 1 (.14) & 1 (.17) & 1.02 & 1.09 \\
Paraphrase (FN) & 1 (.14) & 1 (.17) & 1.02 & 1.09 \\
Abbreviation (FN) & 1 (.14) & 0 (.00) & 1.02 & 0.00 \\
\bottomrule
\end{tabular} 
\end{center}
\caption{Statistics of all FP and FN predictions. Error types are defined in \ref{tab:error_examples_full}. 
Freq refers to the frequency, with relative frequency included in the parenthesis. Error rates (ER) are computed on the corresponding predictions.
}
\label{tab:error_analysis_full}
\end{table}

\section{Annotation }\label{sec:appendix_5}
\begin{figure*}[!ht]
\centering
\includegraphics[width=14cm]{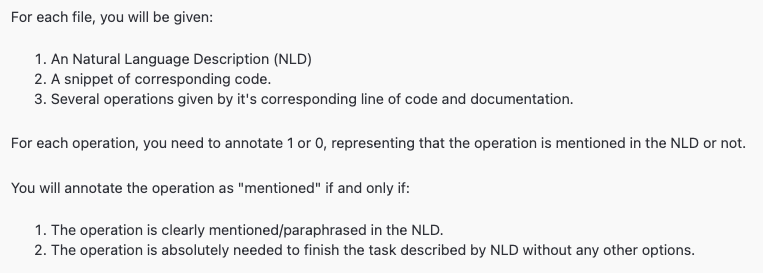}
\caption{The annotation guide.}
\vspace{-0.8em}
\label{fig:annotation_guide}
\end{figure*}

\begin{figure*}[!ht]
\centering
\includegraphics[width=13cm]{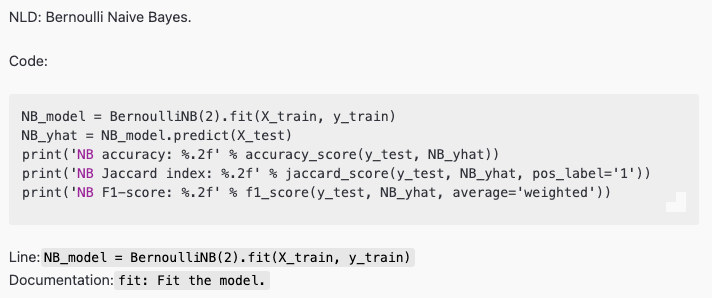}
\caption{An example of annotation. Given an NLD-Code pair and key operations (with documentation) of it, an annotator is required to annotate each key operation as \textit{aligned} or \textit{missing}}
\vspace{-0.8em}
\label{fig:annotation_example}
\end{figure*}

We asked two Ph.D. students to annotate 200 NLD-Code pairs, respectively. It takes a volunteer about 2 hours to annotate. We show the guide in figure~\ref{fig:annotation_guide} and an example of annotation figure~\ref{fig:annotation_example}.

\paragraph{Discrepancy of annotation between development and test set.}\label{paragraph:discrepancy} We noticed the discrepancy of Fleiss Kappa between the development and test set. We then asked annotators to provide reasons for different annotations. As a result, subjectivity is the main reason for differences between annotations. An example is shown in figure~\ref{fig:annotation_example}, where fitting the model is not directly mentioned yet can be inferred from the NLD. We also find that the test set contains more examples like this one, leading to a discrepancy of Fleiss Kappa between the development and the test set. We accept this difference as subjectivity is part of deciding \textit{whether an operation is mentioned}.

\section{Examples of CodeClarQA Dataset}\label{sec:appendix_6}
We present examples from our dataset in Table~\ref{tab:codeclarqa_examples}.

\begin{table*}[!t]
\small
\addtolength{\tabcolsep}{-3pt}
\begin{center}
\begin{tabular}{@{}L{15cm}@{}}
\toprule 
\textbf{NLD}: So, 18 categorical features and 10 numerical features to clean. We start with the numerical features, first thing to do is have a look at them to learn more about their distribution and decide how to clean them. > 2.2 Numerical features.
\\
\textbf{CQ1}: Do you want to call 'pandas.head' documented as 'Return the first `n` rows.'?"
\par
\textbf{A1}: Yes.
\par
\textbf{CQ2}: Do you want to call 'pandas.fillna' documented as 'Fill NA/ NaN values using the specified method.'?
\par
\textbf{A2}: Yes.
\\
\textbf{Code}:
\par
NAnum.head()
\par
c['MasVnrArea'] = c.MasVnrArea.fillna(0)
\par
c['LotFrontage'] = c.LotFrontage.fillna(c.LotFrontage.median())
\par
c['GarageYrBlt'] = c['GarageYrBlt'].fillna(1980)
\\

\midrule[.04em]
\textbf{NLD}: There are many libraries out there that support one-hot encoding but the simplest one is using method. This function is named this way because it creates dummy/indicator variables.
\\
\textbf{No CQAs.}
\\
\textbf{Code}:
\par
Train\_Master = pd.get\_dummies(Train\_Master, columns=['Sex', 'Pclass', 'Embarked'], drop\_first=True)
\par
Train\_Master.drop(['PassengerId', 'Name', 'Ticket'], axis=1, inplace=True)
test\_ids = Test\_Master.loc[:, 'PassengerId']
\par
Test\_Master = pd.get\_dummies(Test\_Master, columns=['Sex', 'Embarked', 'Pclass'], drop\_first=True)
\par
Test\_Master.drop(['PassengerId', 'Name', 'Ticket'], axis=1, inplace=True)
Train\_Master.head()
\\

\midrule[.04em]
\textbf{NLD}: Need to look at the y\_log relationship since that is what we will be predicting in the model.
\\
\textbf{CQ1}: Do you want to call anything related to 'plot'? If yes, which one?
\par
\textbf{A1}: Yes, I want to call 'matplotlib.plot'.
\par
\textbf{CQ2}: Do you want to call 'matplotlib.scatter' documented as 'A scatter plot of *y* vs. *x* with varying marker size and/or color.'?
\par
\textbf{A2}: Yes.
\\
\textbf{Code}:
\par
NAnum.head()
\par
y = np.exp(11.1618915) * np.exp(0.000570762509 * x\_data)
\par
plt.plot(x\_data, np.log(y\_data), 'o')
\par
plt.scatter(x\_data, np.log(y), c='red')
\\

\midrule[.04em]
\textbf{NLD}: Ensembling is a way to increase performance of a model by combining several simple models to create a single powerful model. I will use voting method in this kernal.
\\
\textbf{CQ1}: Do you want to call anything related to 'model/algorithm'? If yes, which one?
\par
\textbf{A1}: Yes, I want to call 'sklearn.RandomForestClassifier'.
\par
\textbf{CQ2}: Do you want to call anything related to 'model/algorithm'? If yes, which one?
\par
\textbf{A2}: Yes, I want to call 'sklearn.LogisticRegression'.
\par
\textbf{CQ3}: Do you want to call anything related to 'model/algorithm'? If yes, which one?
\par
\textbf{A3}: Yes, I want to call 'sklearn.DecisionTreeClassifier'.
\par
\textbf{CQ4}: Do you want to call anything related to 'model/algorithm'? If yes, which one?
\par
\textbf{A4}: Yes, I want to call 'sklearn.GaussianNB'.
\par
\textbf{CQ5}: Do you want to call anything related to 'score'? If yes, which one?
\par
\textbf{A5}: Yes, I want to call 'sklearn.cross\_val\_score'.
\\
\textbf{Code}:
\par
from sklearn.ensemble import VotingClassifier
\par
estimators = [('RFor', RandomForestClassifier(n\_estimators=100, random\_state=0)), ('LR', LogisticRegression(C=0.05, solver='liblinear')), ('DT', DecisionTreeClassifier()), ('NB', GaussianNB())]
\par
ensemble = VotingClassifier(estimators=estimators, voting='soft')
\par
ensemble.fit(train\_X, train\_Y.values.ravel())
\par
print('The accuracy for ensembled model is:', ensemble.score(test\_X, test\_Y))
\par
cross = cross\_val\_score(ensemble, X, Y, cv=10, scoring='accuracy')
\par
print('The cross validated score is', cross.mean())
\\

\toprule
\end{tabular}
\end{center}
\caption{Examples of the CodeClarQA dataset.
}
\label{tab:codeclarqa_examples}
\end{table*}